\documentclass[runningheads]{llncs}

\usepackage[T1]{fontenc}
\usepackage{graphicx}
\usepackage{booktabs}
\usepackage{tabularx}
\usepackage{hyperref}
\usepackage{array}
\usepackage{ragged2e}
\usepackage{amsmath}
\usepackage{listings}
\usepackage{xcolor}
\usepackage{multirow}
\usepackage{amsfonts}

\definecolor{codegreen}{rgb}{0,0.6,0}
\definecolor{codegray}{rgb}{0.5,0.5,0.5}
\definecolor{codepurple}{rgb}{0.58,0,0.82}
\definecolor{backcolour}{rgb}{0.95,0.95,0.92}

\lstdefinestyle{mystyle}{
    backgroundcolor=\color{backcolour},   
    commentstyle=\color{codegreen},
    keywordstyle=\color{magenta},
    numberstyle=\tiny\color{codegray},
    stringstyle=\color{codepurple},
    basicstyle=\ttfamily\footnotesize,
    breakatwhitespace=false,         
    breaklines=true,                 
    captionpos=b,                    
    keepspaces=true,                 
    numbers=left,                    
    numbersep=5pt,                  
    showspaces=false,                
    showstringspaces=false,
    showtabs=false,                  
    tabsize=2
}

\hypersetup{
    colorlinks=true,
    linkcolor=blue,
    filecolor=magenta,      
    urlcolor=blue,
    citecolor=blue,
}

\begin{document}

\title{LLM Essay Scoring Under Holistic and Analytic Rubrics: Prompt Effects and Bias}

\titlerunning{LLM Essay Scoring}
\author{Filip J. Kucia\inst{1}\orcidID{0009-0005-8473-1402} \and
Anirban Chakraborty \inst{2}\orcidID{0000-0001-7425-6664} \and
Anna Wróblewska\inst{1}\orcidID{0000-0002-3407-7570} 
}

\authorrunning{F. Kucia et al.}

\institute{Faculty of Mathematics and Information Science, Warsaw University of Technology,
00-662 Warsaw, Poland\\
\email{\{filip.kucia.dokt,anna.wroblewska1\}@pw.edu.pl}
\and
University of Wolverhampton, UK\\
\email{a.chakraborty@wlv.ac.uk}
}

\maketitle

\begin{abstract}
Despite growing interest in using Large Language Models (LLMs) for educational assessment, it remains unclear how closely they align with human scoring. We present a systematic evaluation of instruction-tuned LLMs across three open essay-scoring datasets (\textbf{ASAP~2.0}, \textbf{ELLIPSE}, and \textbf{DREsS}) that cover both holistic and analytic scoring. We analyze agreement with human consensus scores, directional bias, and the stability of bias estimates. Our results show that strong open-weight models achieve moderate to high agreement with humans on holistic scoring (Quadratic Weighted Kappa $\approx 0.6$), but this does not transfer uniformly to analytic scoring. In particular, we observe large and stable negative directional bias on \textbf{Lower-Order Concern (LOC)} traits, such as Grammar and Conventions, meaning that models often score these traits more harshly than human raters. We also find that concise keyword-based prompts generally outperform longer rubric-style prompts in multi-trait analytic scoring. To quantify the amount of data needed to detect these systematic deviations, we compute the minimum sample size ($N_{\min}$) at which a 95\% bootstrap confidence interval for the mean bias excludes zero. This analysis shows that LOC bias is often detectable with very small validation sets, whereas \textbf{Higher-Order Concern (HOC)} traits typically require much larger samples. These findings support a \textbf{bias-correction-first} deployment strategy: instead of relying on raw zero-shot scores, systematic score offsets can be estimated and corrected using small human-labeled bias-estimation sets, without requiring large-scale fine-tuning.

\end{abstract}

\section{Introduction}
Automated Essay Scoring (AES) remains a challenging open problem in Natural Language Processing (NLP) and educational measurement, driven by the need to scale feedback and assessment in writing-intensive domains. While human annotation is commonly treated as the gold standard, it is resource-intensive, slow, and subject to inter- and intra-rater variability~\cite{shermis2003automated,uto2021item}. Unlike more objective tasks, such as classification or translation, essay evaluation requires assessing open-ended creativity while simultaneously enforcing rigid linguistic constraints, a duality that makes defining a single ``ground truth'' inherently difficult~\cite{beigman2023assessing}. Traditional AES approaches, ranging from surface-level readability metrics (simple text statistics) 
to regression-based neural encoders (e.g., BERT~\cite{devlin-etal-2019-bert}), have achieved partial success, but often struggle to capture high-level rhetorical constructs such as coherence, argumentation quality, and discourse organization~\cite{taghipour2016neural,ramesh2022,ijcai2024p897}. These limitations raise persistent concerns regarding construct validity and generalization beyond narrow training distributions~\cite{williamson2012automated}.

The emergence of large language models (LLMs) has introduced a new paradigm for evaluative NLP. Under the \emph{LLM-as-a-Judge} framework, instruction-tuned models are used to approximate human judgment directly, often in zero-shot or few-shot settings, thereby reducing reliance on task-specific training data~\cite{zheng2023judging,chiang2023large}. Recent studies report that instruction-following LLMs can achieve moderate to high agreement with human raters on holistic evaluation tasks, in some cases approaching human inter-rater reliability~\cite{naismith2023automated}. These findings suggest that LLMs encode substantial implicit knowledge about writing quality and assessment criteria, particularly for holistic judgments of essay quality.

Despite this promise, the reliability and validity of LLM-based judges remain under active scrutiny. Prior work has documented multiple vulnerabilities, including position bias, verbosity bias, sensitivity to prompt phrasing, and instability under superficial changes to the evaluation setup~\cite{wang2023large,kochnavi2024}. More recent analyses further highlight discrepancies between model confidence and actual alignment with human preferences, raising concerns about systematic bias and its calibration (alignment)~\cite{thakur-etal-2025-judging,dietz2025,blatz2023calibration}.
A critical challenge is the \emph{stability} of these evaluators: specifically, how agreement and bias shift across \emph{scoring regimes} -- defined here by the scoring format (holistic vs.\ analytic) and the rubric specification (trait selection and score scale) and how sensitive judgments are to prompt formulation.

In this work, we present a systematic evaluation of instruction-tuned open-weight LLMs as automated essay scorers across multiple datasets and scoring regimes, with a particular focus on the contrast between holistic and analytic evaluation with rubrics that gather predefined traits. Moving beyond simple correlation-based metrics, we adopt a multifaceted analytical framework that integrates agreement analysis, trait-level bias characterization, and our method for determining the minimum sample size required to detect mean bias using bootstrapping, following~\cite{bootstrap}. To analyze performance across rubric traits, we adopt the distinction between \textit{Higher Order Concerns} (HOCs) and \textit{Lower Order Concerns} (LOCs), introduced in writing-center pedagogy as a prioritized order of concerns for revision and feedback~\cite{reigstad1984training}. HOCs capture global, discourse-level properties that shape overall communicative effectiveness (e.g., idea development and organization), whereas LOCs capture local linguistic form, including lexico-grammatical choices and surface correctness (e.g., grammar, usage, and punctuation)~\cite{reigstad1984training}. Accordingly, we treat \textit{Content}, \textit{Organization}, and \textit{Cohesion} as HOC traits, and \textit{Language}, \textit{Syntax}, \textit{Vocabulary}, \textit{Phraseology}, \textit{Grammar}, and \textit{Conventions} as LOC traits. Specifically, we investigate four research questions:

\begin{itemize} 

\item[\textbf{RQ1:}] \textbf{Cross-Dataset Ranking Stability.} 
Holding the scoring regime and prompt strategy constant, are model performance rankings \textbf{stable} across datasets and scoring regimes, despite differences in trait definitions and score scales?
\item[\textbf{RQ2:}] \textbf{Regime-Dependent Agreement.} 
Under a fixed prompt strategy, do models achieve higher agreement with the Human Consensus Score (\emph{HCS}) on holistic scoring tasks than on analytic, multi-trait tasks?
\item[\textbf{RQ3:}] \textbf{Prompt Strategy Sensitivity.} 
How does the prompt strategy (\emph{Keywords} vs.\ \emph{Guidelines}) change agreement and systematic deviations, and is this effect dependent on the scoring regime or trait type?
\item[\textbf{RQ4:}] \textbf{Systematic Scoring Deviations.} 
\begin{itemize}
    \item[\textbf{A:}] Under a fixed prompt strategy, do models exhibit systematic bias or score compression relative to the \emph{HCS}, and does this differ for HOC vs.\ LOC traits?
    \item[\textbf{B:}] If systematic deviations exist, what is the minimum sample size ($N_{\min}$) required to statistically detect them?
\end{itemize}

\end{itemize}

\noindent Our goal is not to introduce a new scoring model, but to clarify the conditions under which LLM-based evaluation can be meaningfully interpreted, adjusted to match human scoring, and validated as a measurement instrument.

\section{Related Work}

Automated Essay Scoring (AES) has been studied extensively for several decades, with early approaches relying on feature-engineering pipelines to predict holistic scores~\cite{williamson2012automated,ramesh2022}. Early systems combined surface-level textual features -- such as essay length, syntactic complexity, and lexical diversity -- with linear or tree-based regression models, as exemplified by the Automated Student Assessment Prize (ASAP) competition~\cite{hewlett2012asap}. While these methods demonstrated that automated scoring could approximate human judgments at scale, subsequent analyses highlighted limitations in construct validity, particularly for higher-level discourse and argumentative features~\cite{shermis2003automated,uto2021item}.

The introduction of neural architectures marked a shift toward representation-based AES. Neural models, including recurrent and transformer-based encoders, reduced reliance on handcrafted features and achieved improved agreement with human raters on benchmark datasets~\cite{taghipour2016neural}. Nevertheless, large-scale reviews of AES research consistently report that even modern neural approaches struggle with interpretability and generalization across prompts and populations~\cite{ramesh2022,ijcai2024p897}. These challenges are particularly pronounced for analytic scoring, where individual rubric dimensions must be assessed independently.

To address these limitations, recent work has emphasized rubric-based and trait-level scoring frameworks. Datasets such as DREsS provide large-scale, expert-annotated benchmarks for evaluating multiple analytic traits in English-as-a-Foreign-Language (EFL) writing~\cite{yoo-etal-2025-dress}. Complementary modeling approaches, such as TRATES, explicitly leverage rubric definitions to guide trait-specific assessment and improve cross-prompt generalization~\cite{eltanbouly-etal-2025-trates}. Taken together, these efforts reflect a growing recognition that analytic scoring requires both fine-grained linguistic sensitivity and alignment with human evaluation practices.

More recently, the emergence of instruction-tuned large language models (LLMs) has given rise to the \emph{LLM-as-a-Judge} paradigm, in which models are prompted to directly evaluate textual outputs without task-specific training~\cite{zheng2023judging}. This approach has been explored across a range of generation tasks, including machine translation, dialogue, and summarization~\cite{fabbri2021summeval,chiang2023large}. Several studies report that LLM-based judges can achieve stronger alignment with human raters when evaluation is framed as a single overall judgment, rather than requiring multiple distinct scoring decisions~\cite{naismith2023automated}.

Despite these promising results, subsequent analyses have identified substantial limitations in the reliability of LLM-based evaluation. Prior work documents sensitivity to prompt formulation, positional and verbosity biases, and instability under superficial changes to evaluation instructions~\cite{wang2023large,kochnavi2024}. More recent studies further emphasize calibration issues, demonstrating that systematic biases may persist even when aggregate agreement appears high~\cite{thakur-etal-2025-judging,dietz2025,blatz2023calibration}. These findings raise concerns about the validity of LLM-based judges as measurement instruments, particularly for analytic evaluation where trait boundaries are less well-defined.

\paragraph{Note on terminology.} Across both AES and LLM-evaluation literature, terminology is used inconsistently. Terms such as \emph{rubric}, \emph{trait}, \emph{dimension}, and \emph{criterion} are often applied interchangeably to denote components of writing quality, while prompt-engineering work may refer to concise labels for these components as \emph{keywords}. In this paper, we use \emph{trait} to denote an individual scoring dimension (e.g., \emph{Cohesion}, \emph{Syntax}) and \emph{rubric} to denote the full set of traits together with their score scales and definitions. Following recent trait-specific rubric-assisted frameworks, we reserve the term \emph{keywords} exclusively for the prompt condition that provides only trait labels, in contrast to the \emph{guidelines} condition that includes full rubric descriptions~\cite{eltanbouly-etal-2025-trates}. Finally, we define the \textbf{Human Consensus Score (HCS)} as the dataset-provided human reference score for each essay. When multiple ratings are available, the benchmark’s specific aggregation procedure is adopted (e.g., averaging or expert adjudication).  We use the HCS as the reference standard (the "ground truth") for evaluating model agreement and systematic scoring deviations, and reserve the term exclusively for the human reference (distinct from model-generated scores).

\section{Methodology}
\label{sec:methodology}

We evaluate instruction-tuned large language models (LLMs) as zero-shot automated essay scorers by treating each model as an independent \emph{rater}. For each essay and trait, the model produces a \textbf{discrete ordinal score} that is compared against the \textbf{Human Consensus Score (HCS)} provided by the benchmark datasets. All results are obtained in a zero-shot setting (no task-specific training or fine-tuning), under controlled prompting conditions.

We quantify three aspects of scoring behavior: (i) \textbf{agreement} with the HCS, (ii) \textbf{systematic scoring deviations} (e.g., harshness/leniency and score variations), and (iii) the \textbf{stability} of these outcomes across prompt strategies and scoring regimes. This framing treats LLMs as \textbf{rater-like scoring systems} for direct comparison to the human reference.

All models are evaluated under identical experimental conditions. The scores are generated independently for each essay and trait using fixed prompt templates (\textit{Keywords} vs.\ \textit{Guidelines}) and deterministic decoding parameters (e.g., temperature $=0$). The analysis spans three benchmark datasets covering both holistic and analytic scoring regimes, allowing us to characterize scoring behavior across rubrics, prompt formulations, and score scales.

\subsection{Datasets}

To capture variation in scoring granularity and rubric structure, we selected three widely used open essay-scoring datasets. Together, they cover holistic assessment and multi-trait analytic evaluation in both native and non-native English writing contexts. Table~\ref{tab:rubric_essay_corpora} provides an overview of these datasets. 

\paragraph{ELLIPSE (Fine-Grained Analytic Scoring).}
The \textbf{ELLIPSE}~\cite{crossley2023english} corpus consists of English Language Learner (ELL) essays scored on a 1.0--5.0 scale (0.5-point increments). The rubric includes six analytic traits: one HOC (\textit{Cohesion}) and five LOCs (\textit{Syntax, Vocabulary, Phraseology, Grammar, Conventions}), enabling comparison of scoring behavior between HOC and LOC traits. 

\paragraph{DREsS (Broad Analytic Scoring).}
The \textbf{DREsS}~\cite{yoo-etal-2025-dress} dataset targets EFL writing, providing scores across three analytic traits: two HOCs (\textit{Content}, \textit{Organization}) and one LOC (\textit{Language}). We evaluate the \texttt{DREsS\_New} and \texttt{DREsS\_Std} subsets, which include student essays assessed by expert raters using standardized rubrics. \texttt{DREsS\_New}\footnote{During preprocessing, 300 entries with missing text were excluded from \texttt{DREsS\_New}, resulting in 1{,}979 essays. \texttt{DREsS\_Std} retained 6{,}508 essays.} contains authentic classroom essays from undergraduate learners, while \texttt{DREsS\_Std} aggregates multiple established AES datasets unified under a common rubric. Synthetic data (\texttt{DREsS\_CASE}) are excluded to focus on authentic student writing. For reporting consistency, we map \texttt{DREsS\_Std} to ``Orig. Train'' and \texttt{DREsS\_New} to ``Orig. Test,'' though both are evaluated strictly in zero-shot settings.

\paragraph{ASAP~2.0 (Holistic Scoring).}
The \textbf{ASAP~2.0}~\cite{hewlett2012asap} dataset serves as a holistic scoring benchmark. It consists of source-based persuasive essays written by U.S. students in grades 6--10. Each essay receives a single integer score on a six-point scale (1--6) representing overall writing quality. Unlike the analytic datasets, this task requires synthesizing multiple quality dimensions into a single scalar judgment.

\begin{table}[t]
\centering
\caption{Overview of essay-scoring datasets used in this study. Essay length statistics are reported in word counts. In \# Essays, the first value is the number used in this study (after preprocessing); the second line reports the original train/test split.}
\label{tab:rubric_essay_corpora}
\scriptsize
\setlength{\tabcolsep}{3.5pt}
\renewcommand{\arraystretch}{1.1}
\begin{tabular}{l c c c c r}
\toprule
\textbf{Dataset} & \textbf{Scoring regime} & \textbf{\# Traits} & \textbf{Assessment scale} & \textbf{\# Essays} & \textbf{Avg.\ Words $\pm$ Std.} \\
\midrule
\href{https://github.com/scrosseye/ASAP_2.0}{ASAP~2.0} & Holistic & 1 & 1--6 (1.0) &
\begin{tabular}[c]{@{}c@{}}24{,}728\\{\tiny 17{,}307/7{,}421}\end{tabular} &
$362.9 \pm 148.5$ \\

\href{https://github.com/scrosseye/ELLIPSE-Corpus}{ELLIPSE} & Analytic & 6 & 1.0--5.0 (0.5) &
\begin{tabular}[c]{@{}c@{}}6{,}482\\{\tiny 3{,}911/2{,}571}\end{tabular} &
$427.8 \pm 191.9$ \\

\href{https://haneul-yoo.github.io/dress/}{DREsS} & Analytic & 3 & 1.0--5.0 (0.5) &
\begin{tabular}[c]{@{}c@{}}8{,}487\\{\tiny 6{,}508/1{,}979}\end{tabular} &
$329.4 \pm 167.6$ \\
\bottomrule
\end{tabular}
\end{table}

\subsection{LLM Scorers}

We evaluate a set of instruction-following open-weight large language models (LLMs) spanning multiple parameter scales and model families to examine the relationship between model capacity and scoring behavior. All models are evaluated in a zero-shot setting using the same prompt strategies (see Section~\ref{sec:ps}) and decoding parameters, with greedy decoding (temperature $=0$) to ensure deterministic outputs for reproducibility.

Our primary focus is on the \textbf{Meta Llama-3.1-Instruct} family, including the \href{https://huggingface.co/meta-llama/Meta-Llama-3.1-8B-Instruct}{8B}, \href{https://huggingface.co/meta-llama/Meta-Llama-3.1-70B-Instruct}{70B}, and  \href{https://huggingface.co/meta-llama/Llama-3.1-405B-Instruct}{405B} parameter variants, which represent a wide range of model capacities within a single architecture. To check whether findings replicate across model families, we additionally evaluate \textbf{GPT-OSS} models with \href{https://huggingface.co/openai/gpt-oss-20b}{20B} and \href{https://huggingface.co/openai/gpt-oss-120b}{120B} parameters. We use all models as released and apply no task-specific fine-tuning on the essay datasets, which allows for a cleaner comparison across scale and model family.

\subsection{Prompting Strategy}\label{sec:ps}
To analyze how the level of instructional detail affects scoring behavior, we designed two prompting strategies for each dataset. The \textbf{\textit{Keywords}} strategy lists only the trait label(s) and scoring scale, whereas the \emph{Guidelines} strategy provides the full rubric text that includes the authors' detailed trait definitions and score-level criteria describing what is required to achieve each score.

\paragraph{Keywords strategy: Trait labels only.}
The prompts in this experiment provide the model only with the names of the trait and the numerical scale:
\begin{lstlisting}
ELLIPSE/DREsS prompt:
    "You are an expert essay grader. Score the essays based on the following rubrics: [List of Traits]. Score Scale: 1.0 (Poor) to 5.0 (Excellent). Use 0.5 increments."
ASAP 2.0 prompt:
    "You are an expert essay grader. Rate the essay on a holistic scale between 1 (minimum) and 6 (maximum). The distance between each grade should be considered equal. Use 1.0 increments."
\end{lstlisting}

\paragraph{Guidelines strategy: Rubric descriptions.}
Prompts in this experiment include comprehensive definitions for every rubric:
\begin{itemize}
\item ELLIPSE: 
    Paragraph-length definitions for levels 1--5 for all 6 traits (e.g., Syntax 5.0: "Flexible and effective use...").
\item DREsS:
    Detailed descriptions for Content ("Paragraph is well-developed..."), Organization ("Argument is very effectively structured..."), and Language ("Sophisticated control...").
\item ASAP 2.0:
    Extensive holistic rating form describing the characteristics of a Score 6 ("clear and consistent mastery", "outstanding critical thinking") down to Score 1 ("severe flaws", "pervasive errors").
\end{itemize}

\subsection{Evaluation Metrics}
We evaluate (i) score \emph{agreement} between the LLM and the Human Consensus Score (HCS) and (ii) the \emph{magnitude and direction} of scoring deviations using the following metrics, computed once again against HCS.\\
The agreement scores are as follows:\\
- \textbf{Quadratic Weighted Kappa (QWK):} Our primary ordinal agreement metric, accounting for the ordinal nature of scores and penalizing larger discrepancies more heavily. (following other studies in AES~\cite{eltanbouly-etal-2025-trates,taghipour2016neural,ijcai2024p897,yoo-etal-2025-dress}).\\
- \textbf{Exact Agreement (EA):} The percentage of instances where the LLM assigns a score numerically identical to the HCS.\\
The magnitude and direction measures are as follows:\\
- \textbf{Bias (Mean Signed Error):} The average signed deviation from the HCS,
    $\hat{\mu}_{\text{bias}}=\frac{1}{n}\sum_{i=1}^{n}\big(score^{(i)}_{\text{LLM}}-score^{(i)}_{\text{HCS}}\big)$,
    where $n$ is the number of essays. Negative values indicate systematic under-scoring (harshness), while positive values indicate over-scoring (leniency).\\
- \textbf{Mean Absolute Error (MAE):} The average absolute difference between LLM and HCS scores, measuring error magnitude regardless of direction.\\
- \textbf{Score Standard Deviation ($\sigma$):} We compute the standard deviation of model-assigned scores ($\sigma_{\text{LLM}}$) and of the HCS ($\sigma_{\text{HCS}}$) separately. Comparing $\sigma_{\text{LLM}}$ and $\sigma_{\text{HCS}}$ is used to detect \emph{score compression} (central tendency/range restriction), where model outputs exhibit reduced score spread relative to the reference.

\noindent Unless stated otherwise, all reported metrics are \textbf{weighted averages} based on the number of essays in each dataset split (Train/Test for ASAP and ELLIPSE; Std/New for DREsS). In this study, we consider an LLM to be well aligned with human raters when it achieves both strong agreement (QWK near 1) and a mean bias close to 0.

\section{Our Results and Analysis}

Our experiments revealed significant performance disparities across datasets and model configurations. Overall, the \emph{Llama-3.1-70B} model using the Keywords prompt strategy emerged as the strongest-performing configuration, consistently achieving the highest QWK agreement and the lowest systematic bias. Conversely, the GPT-OSS-120B model underperformed significantly, showing extreme negative bias.

\subsection{Overall Performance Leaderboard}
\label{sec:leaderboard}
Table~\ref{tab:leaderboard} summarizes performance across models and prompt strategies, highlighting the impact of prompt design on agreement and bias.

\begin{table}[h]
\centering
\caption{Performance of all models by prompt strategy, reported as weighted averages across data splits. \textbf{Bold} values indicate the highest QWK for each model and the signed bias closest to zero for each dataset. Note: All bias estimates are statistically different from zero ($p < 0.001$) according to a Wilcoxon Signed-Rank test.}
\label{tab:leaderboard}
\begin{tabular}{l l cc cc}
\toprule
& & \multicolumn{2}{c}{\textbf{Keywords}} & \multicolumn{2}{c}{\textbf{Guidelines}} \\
\cmidrule(lr){3-4} \cmidrule(lr){5-6}
\textbf{Dataset} & \textbf{Model} & \textbf{QWK} $\uparrow$ & \textbf{Bias} $\to 0$ & \textbf{QWK} $\uparrow$ & \textbf{Bias} $\to 0$\\
\midrule
\multirow{5}{*}{\textbf{ASAP 2.0}} 
 & Llama 3.1 70B  & 0.533 & -0.39 & \textbf{0.601} & \textbf{-0.05} \\
 & Llama 3.1 405B  & 0.496 & -0.62 & \textbf{0.592} & -0.34 \\
 & Llama 3.1 8B  & 0.373 & -0.66 & \textbf{0.463} & -0.44 \\
 & GPT-OSS 120B  & 0.117 & -1.47 & \textbf{0.299} & -0.68 \\
 & GPT-OSS 20B  & 0.137 & -1.45 & \textbf{0.261} & -1.11 \\
\midrule
\multirow{5}{*}{\textbf{ELLIPSE}} 
 & Llama 3.1 70B  & \textbf{0.321} & \textbf{-0.66} & 0.235 & -0.83 \\
 & Llama 3.1 405B & 0.201 & -1.06 & \textbf{0.214} & -1.04 \\
 & Llama 3.1 8B   & \textbf{0.184} & -1.04 & 0.173 & -1.16 \\
 & GPT-OSS 120B   & 0.070 & -1.57 & \textbf{0.100} & -1.49 \\
 & GPT-OSS 20B    & 0.078 & -1.39 & 0.078 & -1.27 \\
\midrule
\multirow{5}{*}{\textbf{DREsS}} 
 & Llama 3.1 70B  & \textbf{0.414} & \textbf{-0.19} & 0.394 & -0.34 \\
 & Llama 3.1 405B & \textbf{0.371} & -0.39 & 0.342 & -0.60 \\
 & Llama 3.1 8B   & \textbf{0.267} & -0.64 & 0.216 & -0.85 \\
 & GPT-OSS 120B   & \textbf{0.091} & -1.33 & 0.074 & -1.49 \\
 & GPT-OSS 20B    & \textbf{0.123} & -1.11 & 0.100 & -1.29 \\
\bottomrule
\end{tabular}
\end{table}

\noindent \textbf{Model performance hierarchy (RQ1).}
Across datasets, the \emph{Llama-3.1-70B} model achieves the strongest alignment with human raters, using HCS as reference, exceeding the substantially larger \emph{Llama-3.1-405B} configuration and GPT-based models. In all 3 datasets, the ranking of the performance models is the same.

\textbf{Impact of scoring granularity (RQ2).}
Performance differs systematically between holistic and analytic evaluation regimes. The highest agreement values are observed on ASAP~2.0, where models assign a single overall quality score (holistic regime). In contrast, analytic scoring is systematically better on DREsS dataset. However, analytic scoring introduces a more difficult setting: on ELLIPSE, which requires simultaneous evaluation of six rubric dimensions, even the strongest configuration reaches only moderate agreement ($0.321$), suggesting that fine-grained diagnostic judgments remain challenging.

\textbf{Effect of prompt strategy (RQ3).}
For holistic scoring (ASAP~2.0), the detailed one-trait description is beneficial: \emph{Guidelines} shows systematic higher agreement than the \emph{Keywords} prompt strategy. Conversely, for analytic scoring (ELLIPSE and DREsS), concise prompts usually yield significantly better agreement: \emph{Keywords} outperforms \emph{Guidelines}.

\subsection{Trait-Level Analysis and Score Distributions}
\label{sec:traits}

Beyond aggregate performance, we examine model behavior at the level of individual rubric dimensions in order to characterize variation in scoring difficulty, score distributions, and systematic deviations from human judgment (HCS). This analysis focuses on the strongest-performing model \emph{Llama-3.1-70B}, see Table~\ref{tab:traits}. 
 
\begin{table}[h]
\centering
\setlength{\tabcolsep}{3.5pt}
\caption{Detailed performance of the top-performing model, \emph{Llama-3.1-70B}, comparing the \textbf{Keywords} and \textbf{Guidelines} prompt strategies.
}
\label{tab:traits}
\resizebox{\textwidth}{!}{%
\begin{tabular}{l l| c | c c c c c | c c c c c}
\toprule
& & \textbf{Ref} & \multicolumn{5}{c|}{\textbf{Keywords Strategy}} & \multicolumn{5}{c}{\textbf{Guidelines Strategy}} \\
\cmidrule(lr){3-3} \cmidrule(lr){4-8} \cmidrule(lr){9-13}
\textbf{Dataset} & \textbf{Trait} & \textbf{$\sigma_{HCS}$} & \textbf{QWK} $\uparrow$ & \textbf{Bias} $\to 0$ & \textbf{MAE} $\downarrow$ & \textbf{EA} $\uparrow$ & \textbf{$\sigma_{\text{LLM}}$} & \textbf{QWK} $\uparrow$ & \textbf{Bias} $\to 0$ & \textbf{MAE} $\downarrow$ & \textbf{EA} $\uparrow$ & \textbf{$\sigma_{\text{LLM}}$} \\
\midrule
\textbf{ASAP 2.0} & Score & 1.01  & 0.533 & -0.39 & 0.73 & 17.4 & 0.72 & \textbf{0.601} & -0.05 & 0.60 & 47.1 & 0.91 \\
\midrule
\multirow{6}{*}{\textbf{ELLIPSE}} 
 & Cohesion (HOC) & 0.66 & \textbf{0.566} & -0.12 & 0.43 & 34.5 & 0.56 & 0.412 & -0.55 & 0.63 & 21.6 & 0.55 \\
 & Syntax (LOC) & 0.66 & \textbf{0.414} & -0.48 & 0.59 & 24.6 & 0.56 & 0.265 & -0.73 & 0.77 & 17.0 & 0.50 \\
 & Vocabulary (LOC) & 0.58 & \textbf{0.278} & -0.55 & 0.64 & 18.6 & 0.47 & 0.236 & -0.67 & 0.71 & 14.9 & 0.41 \\
 & Phraseology (LOC) & 0.66 & \textbf{0.247} & -0.74 & 0.79 & 16.1 & 0.49 & 0.132 & -1.07 & 1.07 & 8.1 & 0.39 \\
 & Conventions (LOC) & 0.67 & \textbf{0.219} & -1.05 & 1.08 & 6.6 & 0.60 & 0.200 & -1.03 & 1.04 & 7.6 & 0.48 \\
 & Grammar (LOC) & 0.69 & \textbf{0.203} & -1.04 & 1.06 & 7.3 & 0.54 & 0.163 & -0.95 & 0.95 & 12.5 & 0.37 \\
\midrule
\multirow{3}{*}{\textbf{DREsS}} 
 & Language (LOC) & 0.79 & \textbf{0.458} & -0.50 & 0.68 & 7.6 & 0.72 & 0.414 & -0.62 & 0.76 & 6.6 & 0.72 \\
 & Organization (HOC) & 0.93 & \textbf{0.393} & -0.24 & 0.73 & 13.3 & 0.76 & 0.362 & -0.40 & 0.77 & 12.4 & 0.74 \\
 & Content (HOC) & 0.99 & 0.391 & 0.18 & 0.73 & 10.9 & 0.70 & \textbf{0.407} & 0.02 & 0.72 & 12.0 & 0.71 \\
\bottomrule
\end{tabular}%
}
\end{table}

Model performance varies across different scoring traits. On the holistic ASAP 2.0 dataset, the model shows high agreement with human raters (\textbf{QWK = 0.601}) and very little scoring bias (\textbf{$-0.05$}). In contrast, analytic scoring shows a clear gap between \textbf{HOC} and \textbf{LOC} traits. On ELLIPSE, the model performs best on the only HOC trait, \textit{Cohesion} (\textbf{QWK = 0.566}). For LOC traits such as \textit{Grammar}, the low agreement (\textbf{QWK = 0.203}) is paired with a large negative bias (\textbf{$-1.04$}). This suggests the model focuses heavily on formal correctness, while human raters are more lenient as long as the essay's message remains clear.

The DREsS dataset shows a different kind of evaluation challenge. Unlike ELLIPSE, where disagreement is often linked to large systematic biases, disagreement on DREsS is primarily caused by high scoring variability. Agreement scores on DREsS are moderate across its traits (e.g., QWK $\approx$ 0.39--0.46). Notably, this moderate agreement is not always paired with a strong directional bias. For example, the \textit{Content} trait has a near-zero scoring bias of \textbf{+0.18}, but the agreement score is still low (\textbf{QWK = 0.391}). This shows that for DREsS, disagreement comes from inconsistent scoring, not from the model being systematically too harsh or too lenient.

Across all datasets, model-assigned scores exhibit consistently lower standard deviations than human annotations for nearly all traits (Table~\ref{tab:traits}). This variance compression limits the model’s ability to differentiate between exceptionally strong and weak essays, providing a plausible explanation for cases where mean bias is small but agreement remains low. Together, these findings underscore the importance of evaluating score distributions alongside average agreement and bias.

\subsection{Systematic Bias and Length Sensitivity}
\label{sec:bias_length}

Beyond standard agreement metrics, it is essential to examine the directionality and validity of model scoring. Systematic deviations from the Human Consensus Score (HCS) can undermine fairness in deployment, while an over-reliance on simple features suggests the model is exploiting spurious correlations - often referred to as shortcut learning - rather than evaluating actual writing quality. To address this, we analyzed mean signed score differences to characterize this \textit{scoring bias}. We also examined the relationship between assigned scores and essay length to investigate whether models exhibit a \textit{verbosity bias}--disproportionately rewarding longer essays--instead 
of engaging with the semantic criteria defined in the rubric.

Across datasets, instruction-tuned models exhibit a tendency toward negative bias, systematically assigning lower scores than human raters. This pattern is especially evident in analytic evaluation settings. On ELLIPSE, weaker-performing configurations such as \emph{Llama-3.1-8B} and \emph{GPT-OSS-120B} display \textbf{severe under-grading}, in some instances \textbf{underestimating scores by more than 1.5 points} on the 5-point scale. Such large discrepancies indicate that, without post-hoc bias correction, these automated systems would substantially undervalue student performance. The \emph{Llama-3.1-70B} model achieves near-zero average deviation on the holistic ASAP~2.0 (Bias: -0.05) and on selected analytic traits in DREsS, suggesting that increased model capacity contributes not only to higher agreement but also to closer alignment with human score distributions.

To assess the potential reliance on essay length as a proxy for quality, we compare the Pearson correlation between scores and word counts ($r$) for both human and model scores. On the holistic ASAP~2.0 dataset, human scores exhibit a strong positive association with essay length ($r_{Human}=0.71$). However, the \emph{Llama-3.1-70B} model shows a notably weaker correlation ($r_{LLM}=0.47$). This significant divergence implies that the model is \textit{less} sensitive to verbosity than human raters in holistic settings. Rather than over-rewarding length, the model appears to adhere more strictly to the rubric's content criteria, avoiding the human tendency to conflate length with quality.

In analytic settings, the behavior varies slightly. On DREsS, the sensitivity is effectively identical ($r_{Human}=0.37$ vs. $r_{LLM}=0.39$). On ELLIPSE, the model displays a moderate increase in length sensitivity compared to humans ($r_{LLM}=0.33$ vs. $r_{Human}=0.18$). However, even this elevated correlation remains far below the threshold of strong association observed in holistic scoring. Taken together, these results indicate that systematic bias in automated scoring arises primarily from differences in scoring strictness (systematic offset), rather than from an exaggerated reliance on essay length.

\subsection{Minimum Sample Size for Bias Detection}
\label{sec:min_bias_detection}

To estimate how much data is needed to detect systematic directional bias (consistent over- or under-grading) in LLM scoring, we compute the minimum sample size ($N_{\min}$) at which the 95\% bootstrap confidence interval for mean bias excludes zero ~\cite{bootstrap,chang2024bootstrap}. For each model--trait--split--strategy combination, we start at $N=5$ and increase the sample size in steps of 5, up to the size of the dataset. At each step, we draw 10{,}000 bootstrap samples (with replacement), compute the mean bias for each resample, and form a 95\% bootstrap percentile confidence interval. If the interval includes zero (we cannot determine if LLM has bias), we increase $N$ and repeat; if it excludes zero, we record that value as $N_{\min}$. If no such $N$ is found within the entire dataset, we report \emph{NR} (not reached).

\paragraph{Results.}
Table~\ref{tab:stability_summary} summarizes the distribution of $N_{\min}$ across datasets. On \textbf{ELLIPSE}, the median $N_{\min}$ is 5 and the 90th percentile is also 5, indicating that directional bias is detectable with very small samples for nearly all model--prompt--trait--split combinations. Combined with the negative mean bias observed in ELLIPSE (approximately $-1.0$), this suggests a highly consistent under-grading pattern.
\begin{table}[t]
\centering
\setlength{\tabcolsep}{3pt}
\renewcommand{\arraystretch}{1.1}

\caption{Summary of minimum sample size for bias detection across datasets. Note: \#Combinations of model–trait–split–strategy.}
\label{tab:stability_summary}

\begin{tabular}{lrrrrrr}
\toprule
\textbf{Dataset} &
\textbf{\#Combinations} &
\textbf{Bias} &
\textbf{Not} &
\textbf{Median} &
\textbf{90th} &
\textbf{Max} \\
& & \textbf{Detected} & \textbf{Reached} &
$\boldsymbol{N_{\min}}$ &
$\boldsymbol{N_{\min}}$ &
$\boldsymbol{N_{\min}}$ \\
\midrule
ASAP~2.0 & 20  & 20  & 0 & 10.0 & 175.5 & 1{,}460.0 \\
ELLIPSE   & 120 & 120 & 0 & 5.0  & 5.0   & 85.0 \\
DREsS    & 60  & 58  & 2 & 5.0  & 176.0 & 650.0 \\
\bottomrule
\end{tabular}
\end{table}
In contrast, \textbf{ASAP~2.0} and \textbf{DREsS} show much broader, highly right-skewed distributions of $N_{\min}$. Although the median remains low ($N_{\min}=10$ for ASAP~2.0 and $N_{\min}=5$ for DREsS), the 90th percentiles rise to approximately 176 essays, and the maximum values exceed 1{,}400. This indicates that some model--prompt--trait--split combinations exhibit detectable directional bias with little data, while others require substantially larger samples because the average bias is small relative to the variability of per-essay biases.

Table~\ref{tab:stability_conditional_70b_final} illustrates this pattern for \emph{Llama-3.1-70B-Instruct}. On the \textbf{ASAP~2.0} holistic task under the Guidelines prompt, detecting non-zero mean bias requires $N=515$ (Orig. Test) and $N=1{,}460$ (Orig. Train), indicating that directional bias is weak relative to the variance of essay-level deviations in this setting. On \textbf{DREsS}, the \textit{Content} trait under the Guidelines prompt is \emph{NR} on the Orig. Train split, meaning that a non-zero mean bias was not detectable under our criterion within the available sample size.

\begin{table}[h]
\centering
\small
\setlength{\tabcolsep}{5pt}
\caption{Minimum sample size for bias detection ($N_{\min}$) for \emph{Llama-3.1-70B-Instruct}, by dataset, trait, prompt strategy, and official split.}
\label{tab:stability_conditional_70b_final}

\begin{tabular}{ll c c c c}
\toprule
\multirow{2}{*}{\textbf{Dataset}} & \multirow{2}{*}{\textbf{Trait}} & \multicolumn{2}{c}{\textbf{Guidelines}} & \multicolumn{2}{c}{\textbf{Keywords}} \\
\cmidrule(lr){3-4} \cmidrule(lr){5-6}
 & & \textbf{Orig. Test} & \textbf{Orig. Train} & \textbf{Orig. Test} & \textbf{Orig. Train} \\
\midrule
\textbf{ASAP 2.0} & Score  & 515 & 1{,}460 & 20 & 20 \\
\midrule
\multirow{6}{*}{\textbf{ELLIPSE}} 
 & Cohesion & 10 & 5 & 70 & 85 \\
 & Conventions & 5 & 5 & 5 & 5 \\
 & Grammar & 5 & 5 & 5 & 5 \\
 & Phraseology & 5 & 5 & 5 & 5 \\
 & Syntax & 5 & 5 & 10 & 10 \\
 & Vocabulary & 5 & 5 & 5 & 10 \\
\midrule
\multirow{3}{*}{\textbf{DREsS}} 
 & Content & 340 & NR & 95 & 100 \\
 & Language & 10 & 5 & 20 & 10 \\
 & Organization & 155 & 15 & 525 & 25 \\
\bottomrule
\end{tabular}
\end{table}

\section{Discussion and Recommendations}
\label{sec:discussion}

Our findings clarify how Large Language Models (LLMs) behave as automated essay scorers across different evaluation settings. For \textbf{cross-dataset stability (RQ1)}, model performance rankings are largely consistent across datasets and scoring regimes. \emph{Llama-3.1-70B-Instruct} remains the strongest overall configuration, while the main rank changes occur among weaker configurations, especially the GPT-OSS models, whose positions swap across datasets but usually with relatively small score differences. This suggests that broad model quality transfers across tasks, even if fine-grained ordering remains sensitive to dataset and prompt conditions (see Table~\ref{tab:leaderboard}).

Building on this, we observe clear \textbf{regime-dependent agreement (RQ2)} between models and human evaluators. Instruction-tuned LLMs show stronger agreement with human raters in holistic scoring (single overall score) than in analytic, multi-trait scoring. On ASAP~2.0, the best configurations track overall writing quality reasonably well, but analytic trait scoring exposes larger deviations. This pattern is especially evident in ELLIPSE, where separate trait scores reveal differences that are less apparent in a single holistic judgment.

We also find strong \textbf{prompt strategy effects (RQ3)}. In analytic scoring, concise keyword-based prompts generally outperform longer rubric-style guidelines. A plausible explanation is contextual interference: long trait descriptions introduce multiple constraints at once, which can reduce consistency across trait judgments and encourage overly strict fallback heuristics. Short keyword prompts, by contrast, appear to provide clearer anchors for trait-specific scoring. In holistic scoring, however, the pattern often reverses, with longer rubric descriptions helping the model form a more coherent overall judgment. Taken together, this shows that prompt effectiveness depends on scoring granularity, not simply on prompt length.

These patterns directly connect to \textbf{systematic scoring deviations (RQ4)}. Across settings, the clearest and most consistent deviation is a negative systematic under-scoring on Lower-Order Concern (LOC) traits, especially Grammar and Conventions, meaning that models tend to score these traits more harshly than human raters. This behavior is better interpreted as a \textit{directional bias} relative to human consensus than as random error or a lack of scoring ability. The minimum-sample-size analysis supports this interpretation: on ELLIPSE, non-zero mean bias is detectable with very small samples for nearly all model--prompt--trait--split combinations, whereas ASAP~2.0 and DREsS show highly right-skewed $N_{\min}$ distributions, including settings where non-zero mean bias is only detectable with large samples or not detectable within the available data. In other words, some configurations show clear directional bias, while others are dominated by the variability of per-essay biases.

\section{Conclusion}

This study presents a systematic evaluation of large language models as automated essay scorers, characterizing their agreement with human raters, systematic biases, and stability across holistic and analytic scoring regimes. The results indicate that strong open-weight models, particularly \emph{Llama-3.1-70B}, achieve substantial agreement with human judgments on holistic assessment tasks (QWK $\approx$ 0.6). However, this alignment does not extend uniformly to analytic evaluation, where stable negative biases emerge on \textbf{Lower Order Concern (LOC) traits} such as Grammar and Conventions.

A key finding of this work is the interaction between prompt formulation and task complexity. For multi-trait analytic scoring, concise \textbf{Keywords}-based prompts generally outperform more detailed \textbf{Guidelines}-based instructions. This suggests that providing extensive rubric text introduces multiple competing constraints simultaneously, particularly when models must apply several evaluative criteria at once. In contrast, compact prompts allow the model to isolate and evaluate individual traits more reliably.

The bootstrap analysis further demonstrates that observed biases are not artifacts of sampling variability but reflect stable model behavior. Biases associated with low-level traits can be detected with relatively small validation sets (often $N \leq 10$), while higher-level semantic traits exhibit greater variance and require substantially larger samples to achieve statistical distinguishability. These results highlight the importance of treating bias estimation as a trait-dependent problem rather than a uniform property of model performance.

Taken together, these findings support a bias-correction-first approach to deploying LLM-based essay-scoring systems. Rather than relying on raw zero-shot scores, practitioners can estimate and correct systematic model deviations using small sets of human-annotated essays. This approach offers a practical alternative to large-scale fine-tuning while preserving alignment with human evaluation standards. More broadly, the study underscores both the promise and the limitations of LLMs as evaluative instruments, particularly for fine-grained diagnostic assessment in educational settings.

\paragraph{Limitations and Future Work}
Our study has several limitations that frame the scope of our findings. First, the evaluation is restricted to \textbf{open-weight, instruction-tuned LLMs in zero-shot settings}. We deliberately excluded proprietary API-only models (e.g., GPT-5, Claude) and fine-tuned systems to focus on reproducible, accessible baselines; however, this means our conclusions regarding calibration and stability may not generalize to closed frontier models or few-shot regimes.

Second, we treat human annotations (HCS) as the gold standard relying on \textbf{single reference score} per essay. Without modeling inter-rater variability, it is difficult to disentangle true model error from legitimate disagreement, particularly for subjective traits like Content or Organization. Future work should incorporate multi-rater benchmarks to better characterize this uncertainty.

Third, our prompting setup employs a single, neutral ``expert grader'' persona and omits \textbf{learner-specific context} (e.g., age, educational level, L1 background). Human raters routinely adjust expectations based on such metadata; the absence of these cues may encourage models to default to an idealized, native-speaker standard, potentially explaining the observed strictness on mechanical traits. We also note that we evaluate models as static measurement instruments, without exploring \textbf{human-in-the-loop feedback} or iterative alignment strategies that would characterize real-world deployment.

Finally, while we quantify bias stability, we do not investigate the \textbf{causes} of the observed strictness (e.g., training data artifacts vs.\ alignment tuning). Future research should explore whether structured elicitation methods, such as requiring evidence-based justification before scoring, can mitigate these biases and improve alignment across fine-grained analytic dimensions.

\begin{credits}
\subsubsection{\ackname}
All authors were funded by the European Union under the Horizon Europe project OMINO (grant agreement No.~101086321). Views and opinions expressed are, however, those of the authors only and do not necessarily reflect those of the European Union or the European Research Executive Agency. Neither the European Union nor the European Research Executive Agency can be held responsible for them. A.W. and F.J.K. were also co-financed by the Polish Ministry of Education and Science under the programme ``International Co-Financed Projects''.
\end{credits}
\bibliographystyle{splncs04}
\bibliography{references}

\end{document}